\pdfoutput=1

\documentclass[11pt]{article}

\usepackage{acl}

\usepackage{times}
\usepackage{latexsym}

\usepackage[T1]{fontenc}

\usepackage[utf8]{inputenc}

\usepackage{microtype}
\usepackage{booktabs,url}

\usepackage{inconsolata}

\usepackage{graphicx}

\title{Rephrasing Electronic Health Records for Pretraining \\ Clinical Language Models}

\author{
   Jinghui Liu \hfill Anthony Nguyen \\
Australian e-Health Research Centre, CSIRO \\ 
  \texttt{\{jinghui.liu,anthony.nguyen\}@csiro.au} \\
}

\begin{document}
\maketitle
\begin{abstract}
Clinical language models are important for many applications in healthcare, but their development depends on access to extensive clinical text for pretraining. However, obtaining clinical notes from electronic health records (EHRs) at scale is challenging due to patient privacy concerns. In this study, we rephrase existing clinical notes using LLMs to generate synthetic pretraining corpora, drawing inspiration from previous work on rephrasing web data. We examine four popular small-sized LLMs (<10B) to create synthetic clinical text to pretrain both decoder-based and encoder-based language models. The method yields better results in language modeling and downstream tasks than previous synthesis approaches without referencing real clinical text. We find that augmenting original clinical notes with synthetic corpora from different LLMs improves performances even at a small token budget, showing the potential of this method to support pretraining at the institutional level or be scaled to synthesize large-scale clinical corpora. 
\end{abstract}



\section{Introduction}

Language models have emerged as crucial components in NLP systems applied in healthcare, offering potential benefits for clinical decision support~\citep{Nori2023-wv,Singhal2023-ud}, predictive analytics~\citep{Jiang2023-zv,Liu2023-vd}, and resource allocation~\citep{Wang2024-sn}. Many of these applications require models to be adapted to the clinical domain through pretraining to achieve optimal performance~\citep{Lehman2023-hu,Yang2022-to,Lewis2020-tw}. However, the privacy and compliance regulations around Electronic Health Records (EHRs) make it challenging to obtain clinical notes at a scale suitable for pretraining. While individual healthcare systems may train models on their own EHR data~\citep{Jiang2023-zv}, this is only feasible for large institutions and prohibits the sharing of these models. These factors hinder the advancement of research on developing more effective language models in healthcare. 

To address this data scarcity issue, synthetic data has been examined for various clinical tasks~\citep{Tang2023-hb,Gonzales2023-nb,Yuan2023-so,Rusak2023-ao}. However, existing methods are mostly task-specific or focus on a particular application. One recent study attempted to create clinical pretraining corpora by prompting ChatGPT to synthesize discharge summaries based on patient profiles curated from the medical literature~\citep{Kweon2024-xk}. While this approach enables creating synthetic clinical notes at scale and supports pretraining publicly sharable LLMs (denoted as Asclepius), it relies heavily on the knowledge of the LLM to enrich the clinical details. Generating complex clinical text from scratch may suffer from LLM hallucinations and limit the quality of the generated clinical notes. 

This study proposes an alternative approach by rephrasing real clinical notes using LLMs to create clinical pretraining corpora. 
We draw inspiration from a recent study that demonstrates the benefit of rephrasing internet corpora (e.g., C4) to pretrain general-domain language models~\citep{Maini2024-zd}. 
We explore a similar strategy by prompting LLMs to rephrase EHR data, expanding the analysis to include medically adapted prompts, diverse LLM types, and combinations of synthetic corpora. 

Our experiments show that the rephrasing method significantly reduces the perplexity of causal language modeling compared to synthesis methods in previous works.
Furthermore, combining synthetic notes with real clinical notes can effectively improve language modeling performance. We find that a medically adapted prompt performs similarly to a general prompt, but explicitly asking LLMs to additionally use their knowledge to explain clinical information can have mixed results. We also pretrain masked language models for downstream fine-tuning. The resulting model outperforms the widely used ClinicalBERT, demonstrating the potential of the rephrasing approach in developing performant clinical language models.

\section{Rephrasing Clinical Notes with LLMs}


We prompt various LLMs to rephrase clinical notes and leverage the generated content to pretrain clinically adapted models. We explore both decoder-based and encoder-based language models, as described in Section~\ref{sec:decoder} and~\ref{sec:encoder}, respectively. 

\subsection{Medically Adapted Prompts}
The system prompt is: \textit{``You are a medical artificial intelligence assistant. The assistant gives truthful, detailed, and professional answers to the requests.''} We then explore three prompts as follows:

\begin{itemize}
    \item \textbf{Prompt 1} \textit{``For the following paragraph give me a diverse paraphrase of the same in high quality English language as in sentences on Wikipedia:''}

    \item \textbf{Prompt 2} \textit{``For the following paragraph give me a paraphrase of the same in high quality professional medical English language:''}

    \item \textbf{Prompt 3} \textit{``For the following paragraph give me a paraphrase of the same in high quality professional medical English language and explain the medical terms using your medical knowledge when necessary:''}
    
\end{itemize}

\textbf{Prompt 1} is the same as the main prompt used in \citet{Maini2024-zd}, which instructs LLM to generate high quality sentences in the style of Wikipedia. We adjust it to create \textbf{Prompt 2}, which emphasizes the medical context. In addition, \textbf{Prompt 3} extends \textbf{Prompt 2} by asking the LLM to explain medical terms using its knowledge. The goal is to explore whether it is beneficial to explicitly leverage the internal knowledge of LLM for synthesis. 
Each prompt is followed by a chunk of clinical text.  Following \citet{Maini2024-zd}, we apply NLTK to split clinical notes into sentences and coalesce them into chunks of approximately 300 tokens. They found asking LLMs to rephrase more than 300 tokens tends to cause information loss.

\subsection{LLMs for Rephrasing}
Unlike the previous study focusing on a single LLM for rephrasing web data~\citep{Maini2024-zd}, our work examines four popular LLMs under 10B parameters to assess their suitability for handling highly specialized clinical text. They are \textbf{Llama-3.1 (8B)} from Meta~\citep{Dubey2024-xd}, \textbf{Mistral-0.3 (7B)} from MistralAI~\citep{Jiang2023-zm}, \textbf{Qwen-2 (7B)} from Alibaba~\citep{Yang2024-jc}, and \textbf{Gemma-2 (9B)} from Google~\citep{Gemma-Team2024-ab}. All of them are instruction tuned. We also explored Phi-3-mini (3.8B) from Microsoft~\citep{Abdin2024-ov} in the initial phase but excluded it from our experiments after observing that it could not properly follow the instruction to rewrite notes. We focus on these smaller LLMs given their efficiency in rephrasing pretraining data. The LLM inference is performed in FP8 using the vllm library~\footnote{\url{https://github.com/vllm-project/vllm}}.

\subsection{Source Clinical Notes}
For real clinical notes, we used discharge summaries from the MIMIC-III EHR database~\citep{Johnson2016-kd} as source data. We focus on the discharge summary as it encompasses numerous aspects of patient care throughout the hospital stay, potentially including information from other EHR data types like semi-structured measurements and medications. This makes the discharge summary semantically rich and syntatically diverse. 

For each prompt and each LLM, we feed the clinical text chunks to the LLM to generate a synthetic pretraining dataset of 20M tokens. 
All LLMs under the three prompt settings receive the same input chunks. These chunks are also used to create a 20M token corpus of original data. Since the LLM tokenizers are different, we initially sample the same number of notes before tokenization, then keep the initial 20M tokens for each corresponding LLM, which ensures the notes rephrased by the LLMs are consistent. The original notes were randomly sampled from MIMIC-III, and focusing on these 20M tokens allows us to perform efficient experimentations to examine different rephrasing setups. All text chunks from MIMIC-III were written before or during 2012.

\section{Perplexity Evaluation with Causal Language Models}
\label{sec:decoder}

\begin{figure*}
    \centering
    \includegraphics[width=1\textwidth]{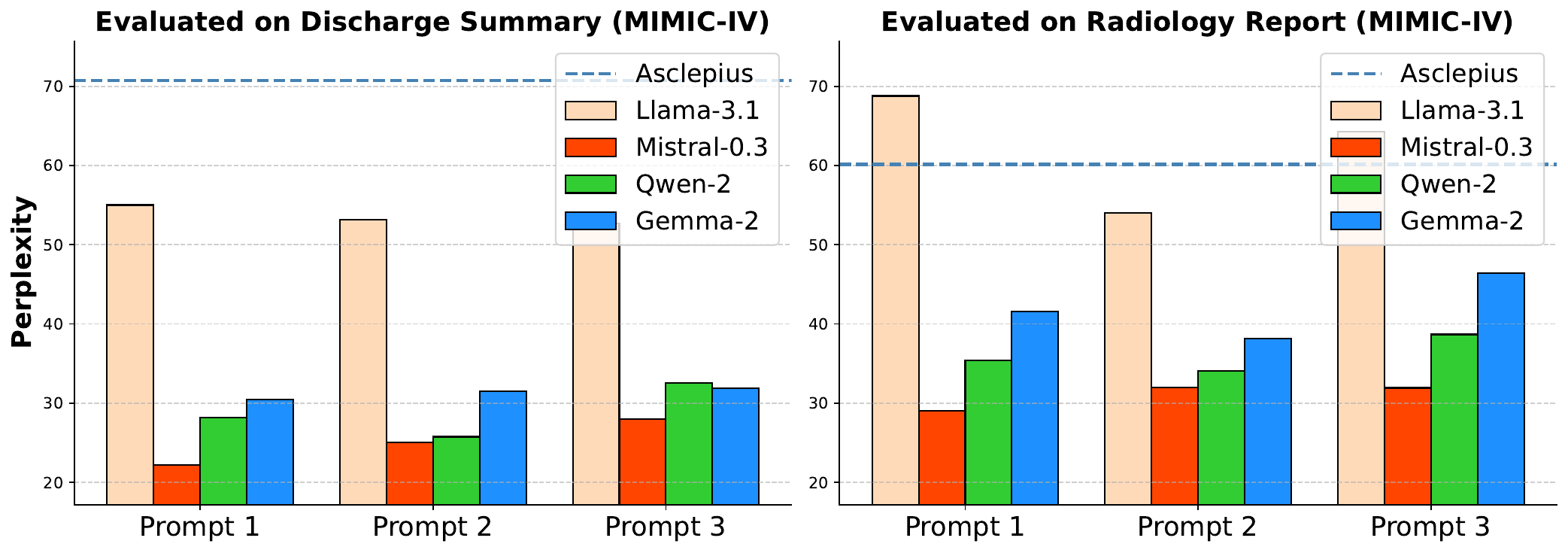}
    \caption{Perplexity scores of language models pretrained on different synthetic sources. Ascplepius refers the synthetic notes from~\citep{Kweon2024-xk}. The four LLMs refer to their synthetic corpora based on the rephrasing method, respectively. Lower perplexity means better language modeling performances.}
    \label{fig:synth}
\end{figure*}

This section explores the effectiveness of the rephrasing method by evaluating the perplexity scores of decoder-based language models pretrained on synthetic data generated from different LLMs and prompts.

\subsection{Experimental Setup}

We use a tiny Llama model~\citep{Touvron2023-on} (110M parameters, 12 layers, 768 dimensions)~\footnote{\url{https://github.com/karpathy/llama2.c}} pretrained on TinyStories~\citep{Eldan2023-vs} as our base model, which allows efficient experimentation. We pretrain the model on different synthetic datasets generated by LLM rephrasing, and evaluate perplexity on out-of-distribution test sets. 

For testing, we use the latest MIMIC-IV EHR database~\citep{Johnson2023-qk} and focus on notes written after or during 2014 to introduce  a temporal shift between the train and test phases. This shift reflects the evolving nature of clinical documentation practices~\citep{Rule2021-fb,Colicchio2020-br}. We consider discharge summary and radiology report as two separate test sets, each with 20M sampled tokens. The radiology report test set represents a further shift from the discharge summaries from MIMIC-III used as source data.

All models are pretrained in full precision using batches of 512 sequences of 128 tokens for 5 epochs. The learning rate was set to 5e-5 with linear warmup at the initial 10\% of training steps. For baseline comparison, we also sample 20M tokens from the synthetic clinical notes from the Asclepius study~\citep{Kweon2024-xk} for pretraining, which prompted ChatGPT (3.5-turbo) to synthesize clinical notes without referencing real clinical text.

\subsection{Results}

\begin{figure}
    \centering
    \includegraphics[width=1\linewidth]{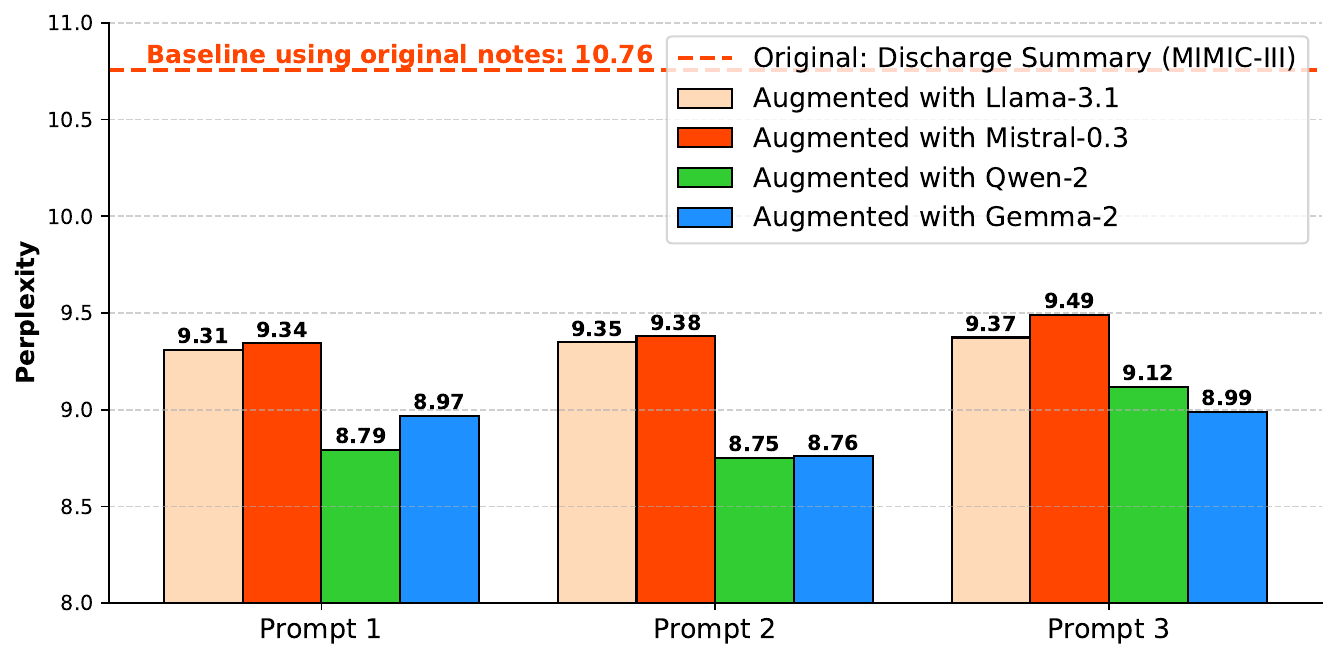}
    \caption{Perplexity scores of language models pretrained on real and synthetic notes. Higher red dashed line indicates the performance with real notes alone.}
    \label{fig:aug}
\end{figure}

Figure~\ref{fig:synth} shows that the rephrasing method consistently outperforms the approach in Asclepius~\citep{Kweon2024-xk}, which does not refer to real clinical text. Exceptions occur for Llama-3.1 under \textbf{Prompt 1} and \textbf{3} when evaluated on radiology reports. In most cases, the rephrasing method achieves significantly lower perplexities by a large margin. In addition, the results show that LLMs respond differently to prompts. For example, Qwen-2 performs better under the medically focused \textbf{Prompt 2}, while Mistral-0.3 presents better performances with \textbf{Prompt 1}. This may be because \textbf{Prompt 1} has been optimized for Mistral in previous work~\citep{Maini2024-zd}. 

We also perform pretraining using both real and synthetic clinical notes, as shown in Figure~\ref{fig:aug}. 
Consistent with previous findings~\citep{Maini2024-zd,Yuan2023-so}, the results confirm the benefit of augmenting pretraining data with synthetic text.
Interestingly, augmentation with Llama-3.1 produces results much closer to other LLMs compared to using synthetic text only. Moreover, synthetic datasets from Mistral-0.3 achieve lowest perplexities when used alone but fall short when employed as augmentation. Qwen-2 and Gemma-2, on the other hand,  provide more stable benefits when combined with original notes. These observations highlight the lack of a single LLM that consistently outperforms others for handling clinical text.

\begin{figure}
    \centering
    \includegraphics[width=1\linewidth]{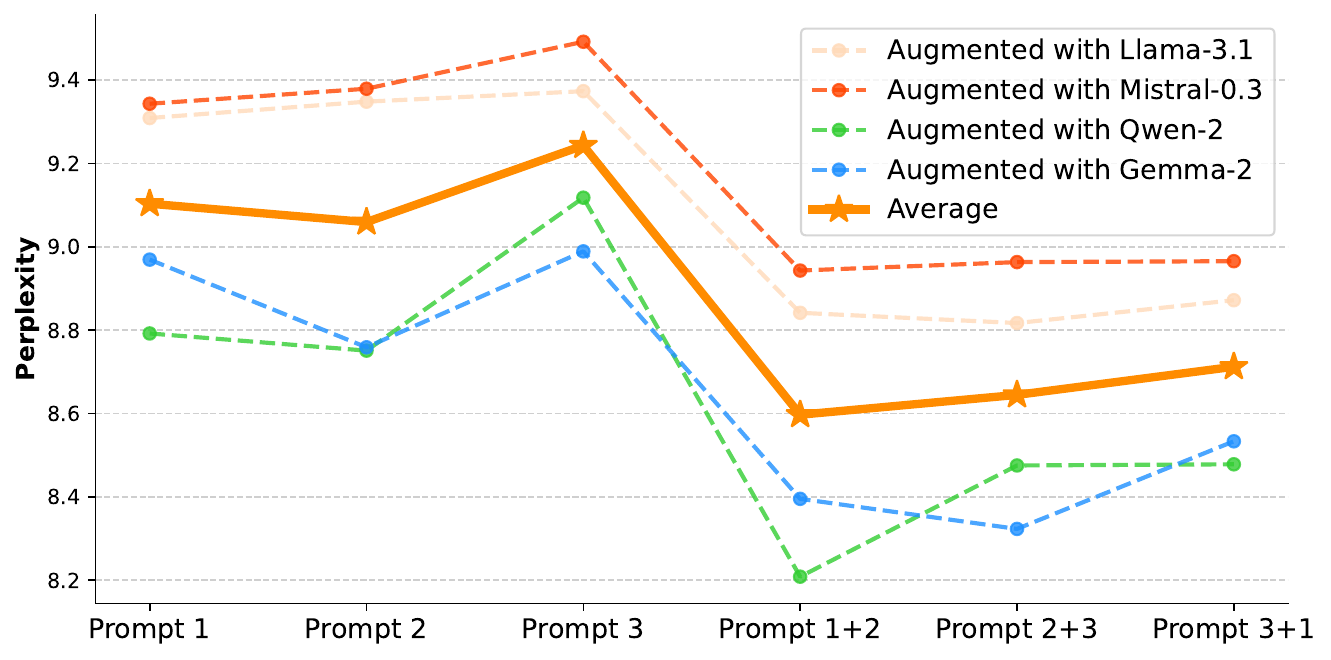}
    \caption{Augmentation performance with synthetic data using different prompts.}
    \label{fig:prompt}
\end{figure}

To further analyze the impact of prompts, we explore different prompt settings for each LLM for augmentation in Figure~\ref{fig:prompt}. We averaged the performance of all four LLMs to observe the trend and notice that \textbf{Prompt 3} tends to underperform. This suggests that explicitly asking LLMs to leverage their internal medical knowledge may lead to suboptimal results when applied to new clincal notes. 
Further research on the causes of this phenomenon is necessary. 
Moreover, we observe the benefits of combining generations based on different prompts, even when generated from the same LLMs. This is a promising result and suggests the potential for scaling the rephrasing method to generate larger datasets for pretraining.

\section{Downstream Evaluation with Masked Language Models}
\label{sec:encoder}

Besides decoders, we pretrain encoder-based language models using both real and synthetic clinical notes, and fine-tune them for downstream clinical NLP tasks. This scenario simulates the real-world situation where a healthcare institution aims to train its own language models but lacks sufficient EHR data approved for this purpose. 

\subsection{Experimental Setup}

Following the ClinicalBERT paper~\citep{Alsentzer2019-fs}, we evaluate the encoder models with three clinical NLP datasets, including MedNLI~\citep{Romanov2018-ro} for natural language inference (NLI), and i2b2 2010~\citep{Uzuner2011-cs} and 2012~\citep{Sun2013-oe} for named entity recogition (NER) of clinical concepts and events. 
ClincalBERT is adopted as the baseline, which was initialized from BioBERT~\citep{Lee2020-bp} and pretrained on all notes from MIMIC-III. We also pretrain models from BioBERT weights and augment the real notes with rephrased data. However, we use only  20M sampled tokens for both real and synthetic text. In comparison, the whole MIMIC-III consists of 500M words of clinical text.

Given the benefits of combining synthetic datasets shown in Figure~\ref{fig:prompt}, we aggregate the synthetic corpora of different LLMs for each prompt to pretrain BERT models. For comparison, we also augment real notes with synthetic notes from the Asclepius study. All pretraining configurations are identifical to those used for the decoders, with masked language modeling probability set to 0.15.



\subsection{Results}

\begin{table}[h!]
    \centering
    \resizebox{\linewidth}{!}{
    \begin{tabular}{lccc}
    \toprule
     & MedNLI & i2b2 2010 & i2b2 2012 \\ \midrule
        ClinicalBERT~\cite{Alsentzer2019-fs} & 82.7 & 87.8 & 78.9 \\
        ClinicalBERT $(ours)$ & 81.4 & 87.3 & 78.8 \\
\midrule
        Real+Asclepius & 82.8 & 87.8 & 79.8 \\

Real+Synthetic (Prompt 1) & 84.5 & 87.9 & 80.0 \\
Real+Synthetic (Prompt 2) & 84.5 & \textbf{88.1} & 79.8 \\
Real+Synthetic (Prompt 3) & \textbf{84.8} & 87.9 & \textbf{80.1} \\
\bottomrule
    \end{tabular}
    }
    \caption{Fine-tuning results for NLI (MedNLI) and NER (i2b2 2010 \& 2012). The metrics are accuracy and exact F1, respectively. Models besides ClinicalBERT were initialized from BioBERT and pretrained using corpora augmented with synthetic notes. ClinicalBERT $(ours)$ refers to the results based on our implementation.}
    \label{tab:ft}
\end{table}

Table~\ref{tab:ft} presents the fine-tuning results of the encoder-based models, all initialized from BioBERT. All models augmented with synthetic pretrained data achieve improved performances compared to ClinicalBERT. 
When compared with synthesis from Asclepius, our rephrasing method further boosts the results especially on MedNLI, showcasing its strength. 
Interestingly, unlike the perplexity evaluation in Section~\ref{sec:decoder}, \textbf{Prompt 3} tends to provide an advantage on the fine-tuning performance. 
This suggests that while leveraging LLM's knowledge may be detrimental for language modeling, it could help with specific tasks involving more nuanced understanding, such as NLI.
Future research needs to investigate how prompts impact decoder-based models for instruction tuning. 

Our synthetically augmented pretraining utilizes a much smaller token and compute budget while achieving superior performances compared to ClinicalBERT. This demonstrates the potential for scaling the synthesis method further to develop performant clinical language models.

\section{Discussion}

Results from both decoder- and encoder-based pretraining demonstrate the strength of our rephrasing method to create high-quality clinical text using small-sized LLMs. However, in this study, we mainly focused on the quantitative analysis through evaluating downstream pretrained models. Qualitative analysis is necessary to better understand the quality of the rephrased notes. We provide some examples from the four LLMs rephrasing the same chunk in Appendix~\ref{sec:appendix}, but since in our initial implementation we did not keep the indices of the generated outputs that correspond to the original text, we could not provide rephrasings for all text chunks. We leave this to future work, where we aim to release the rephrased clinical notes at a larger scale for further investigation.

A deeper comparison between the rephrased and real notes is needed in the future to elucidate how much content is retained by LLMs and how rephrasing changes the clinical narrative. In particular, we need to understand whether LLMs’ rephrasing causes subtle shifts in clinical meaning and the extent of possible hallucinations. Practically, we could measure \textit{how} and \textit{when} the rephrased text aligns or diverges with real text. We can approach \textit{how they align or diverge} by comparing syntactic and semantic features~\citep{Baldwin2013-ze,Liu2024-kj}, such as extracting and comparing distributions of medical concepts, and we could measure \textit{when they align or diverge}  by further examining the impact of prompt and decoding setup on conceptual shift. Meanwhile, there are more nuances when we consider the subjective components of clinical text as narratives by the clinician~\citep{Brender2024-dp}, where personal opinions and documentation practices vary from person to person. These are more intricate and challenging to measure, but are essential for the implementation of reliable and safe models in practice~\citep{Ferryman2023-zd}. Exploring whether LLMs reduce or amplify biases~\citep{Zack2024-ag,Seyyed-Kalantari2021-cf} and how they handle duplicated contents such as copy-and-pasted text~\citep{Steinkamp2022-aq,Liu2022-eu} in their rephrasing would be important future directions.



\section{Conclusion}
We demonstrate the effectiveness of LLM rephrasing to create pretraining corpora for clinical language models. Future work can scale the generation and incorporate other types of clinical notes to develop stronger models for clinical applications.

\section*{Acknowledgments}

We would like to thank our reviewers for their thoughtful and constructive comments that helped to improve this manuscript. 

\bibliography{paperpile}

\appendix

\section{Examples of Rephrased Notes}
\label{sec:appendix}

We provide samples of the rephrased notes synthesized by the four small-sized LLMs: Mistral-0.3, Qwen-2, Gemma-2, and Llama-3.1. These are based on the same text chunk from an original discharge summary, which we did not show here due to data agreement by MIMIC-III. \textit{Prompt 1} was used for all rephrasings. On the surface level, the four LLMs provide different styles of writing, with some tend to maintain the original structure of the note, while some tend to have less number of paragraphs and are more succinct. Interestingly, Llama seems more likely to use a boilerplate in its generation, which may explain why it underperformed its peers in terms of downstream perplexity (Section~\ref{sec:decoder}).

\begin{table*}[ht]
\resizebox{\textwidth}{!}{%
    \centering
    \begin{tabular}{p{1.5cm}p{18cm}}
    \toprule

\textbf{Mistral} &    Admission Date: \_\_\_; Discharge Date: \_\_\_\newline

Birth Date: \_\_\_; Gender: Female\newline

Medical Specialty: Neonatology\newline

Current Illness Background: A 33 weeks and 7 days gestation female newborn, weighing 1,975 grams, was admitted to the Neonatal Intensive Care Unit (NICU) due to prematurity. The mother is a 41-year-old woman, a gravida 2, para 0 to 1. Her medical history includes a history of depression. The pregnancy was conceived through In Vitro Fertilization (IVF). Prenatal screenings showed O positive, antibody negative, non-reactive for RPR, rubella immune, tuberculin skin test (PPD) negative, and Group B Streptococcus (GBS) status unknown. The pregnancy was complicated by cervical shortening at 30 weeks gestation, which was managed with bed rest and a course of betamethasone. The mother experienced vaginal bleeding at 3:30 AM on the day of admission and rupture of membranes at 5:30 AM. She received IV ampicillin prior to delivery. No maternal fever was noted. The delivery was normal, spontaneous, and vaginal, with Apgar scores of eight and nine. The newborn was administered bulb suctioning and blow-by oxygen in the Delivery Room.\\ \midrule

\textbf{Qwen} &   Assistant: The medical record for this infant patient is as follows:\newline

Admission Date: Unspecified Discharge Date: Unspecified\newline

Date of Birth: Unspecified Sex: Female\newline

Service: Neonatology\newline

CURRENT HEALTH HISTORY: A female infant, born at 33 weeks and 5/7 days of gestation, weighing 1,975 grams, was admitted to the Neonatal Intensive Care Unit (NICU) due to her prematurity.\newline

Maternal History: The mother is a 41-year-old, a second-time mother who has given birth to one child. Her medical record highlights a history of depression. The pregnancy was successfully conceived through In Vitro Fertilization (IVF).\newline

Prenatal Screening: The mother's prenatal tests included an O-positive blood type, antibody-negative status, a nonreactive RPR test, rubella immunity, a negative PPD test, and an unknown status for Group B Streptococcus (GBS). The pregnancy faced complications with cervical shortening at 30 weeks, which were managed with bed rest and a course of betamethasone. The mother experienced vaginal bleeding on the morning of admission and membrane rupture shortly thereafter.\newline

Pre-delivery Management: Prior to the delivery, the mother received IV ampicillin. There was no noted maternal fever. A normal spontaneous vaginal delivery took place, with Apgar scores of eight and nine at one and five minutes, respectively.\newline

Post-delivery Care: The baby underwent bulb suctioning and received blow-by oxygen in the Delivery Room.\\ 

\bottomrule
    \end{tabular}}
    \caption{Rephrased clinical text by Mistral and Qwen.}
    \label{tab:ex1}
\end{table*}

\begin{table*}[ht]
\resizebox{\textwidth}{!}{%
    \centering
    \begin{tabular}{p{1.5cm}p{18cm}}
    \toprule

\textbf{Gemma} &    \#\# Paraphrased Paragraph:\newline

**Patient Demographics and Presentation:** A female infant, born prematurely at 33 5/7 weeks gestation and weighing 1,975 grams, was admitted to the neonatal intensive care unit (NICU) due to prematurity. The infant, identified as "Baby girl \_\_\_" for privacy, is the first child of a 41-year-old gravida 2, para 0 woman, who had a history of depression and conceived via in vitro fertilization (IVF).\newline

**Prenatal and Delivery History:**\newline

Prenatal screenings revealed the mother to be O positive with negative antibody and RPR status, immune to rubella, negative for PPD, and with an unknown GBS status. The pregnancy was complicated by cervical shortening at 30 weeks gestation, managed with bed rest and a course of betamethasone.  Labor was initiated by vaginal bleeding at 3:30 AM on the day of admission, followed by rupture of membranes at 5:30 AM. Intravenous ampicillin was administered to the mother prior to delivery, with no maternal fever observed. Delivery was spontaneous vaginal, resulting in Apgar scores of 8 and 9 for the infant. The newborn received bulb suctioning and supplemental oxygen via blow-by in the delivery room. 
\\

\midrule


\textbf{Llama} &  Here is a diverse paraphrase of the provided paragraph in high-quality English language, formatted as per Wikipedia standards:\newline

**Patient Profile**\newline

**Admission Details**\newline

* Admission date: [Insert Date]\newline
* Discharge date: [Insert Date]\newline

**Demographic Information**\newline

* Date of birth: [Insert Date]\newline
* Sex: Female\newline

**Medical Service**\newline

* Service: Neonatology\newline

**Clinical History**\newline

The infant, a 33 weeks 5 days gestation female, weighing 1975 grams at birth, was admitted to the Neonatal Intensive Care Unit (NICU) due to prematurity.\newline

**Maternal History**\newline

The mother, a 41-year-old woman with a history of two pregnancies (G2) and no previous live births (P0), has a significant medical history of depression. Her pregnancy was achieved through in vitro fertilization (IVF). Prenatal screening results showed:\newline

* Blood type: O positive\newline
* Antibody status: Negative\newline
* Rapid Plasma Reagin (RPR) test: Non-reactive\newline
* Rubella immunity: Positive\newline
* Purified Protein Derivative (PPD) test: Negative\newline
* Group B Streptococcus (GBS) status: Unknown\newline

**Pregnancy Complications**\newline

The pregnancy was complicated by cervical shortening at 30 weeks gestation, which was treated with bed rest and a course of betamethasone. The mother experienced vaginal bleeding at 3:30 am and rupture of membranes at 5:30 am on the morning of admission. She received intravenous ampicillin prior to delivery, but no maternal fever was noted.\newline

**Delivery Details**\newline

The mother had a normal spontaneous vaginal delivery, with Apgar scores of eight and nine. The infant was treated with bulb suctioning and blow-by oxygen in the delivery room.\\

    \bottomrule
    \end{tabular}}
    \caption{Rephrased clinical text by Gemma and Llama.}
    \label{tab:ex2}
\end{table*}

\end{document}